\documentclass{article} 
\usepackage{iclr2027_conference,times}

\usepackage{amsmath,amssymb}
\usepackage{graphicx}
\usepackage{booktabs}
\usepackage{multirow}
\usepackage{float}
\usepackage{microtype}
\usepackage[hidelinks]{hyperref}
\usepackage{url}

\newcommand{\sig}{\sigma}
\newcommand{\gpt}{\textsc{GPTQ}}
\newcommand{\rtn}{\textsc{INT4-RTN}}
\newcommand{\nf}{\textsc{NF4}}

\title{The Vulnerability Window: Decision Instability \\ in Quantized Reward Models}

\author{Omkar Shewale \\
Illinois Institute of Technology \\
\texttt{oshewale@hawk.illinoistech.edu}
\And
Deepak Kumar \\
Illinois Institute of Technology \\
\texttt{dkumar15@hawk.illinoistech.edu}
\And
Divakar Kumar Yadav \\
University of Wisconsin--Milwaukee \\
\texttt{dkyadav@uwm.edu}}

\newif\ifreviewcopy
\reviewcopytrue

\ifreviewcopy\iclrfinalcopy\fi

\begin{document}

\maketitle
\ifreviewcopy
  \lhead{\small\itshape The Vulnerability Window}
  \renewcommand{\headrulewidth}{0pt}
\fi

\begin{abstract}
Reward models and LLM judges are routinely quantized to 4 bits so that scoring many candidates
stays affordable, and the usual check is that benchmark accuracy barely moves. That check is close
to blind. Across eight reward models from two families ($0.6$B--$8$B), three generative judges and
four 4-bit quantizers, quantization rewrites $1.1$--$12.3\%$ of pairwise preference decisions on
RewardBench while aggregate accuracy moves by as little as $0.03$ points ($p=1.00$), because the
errors it fixes and introduces cancel. Flips concentrate where the full-precision margin is small
relative to the quantization noise scale $\sig$: a Gaussian approximation $\Phi(-|m|/\sig)$
describes the \emph{aggregate} rate, over-predicting it in all 30 scalar settings and failing for
generative judges, and occupancy of the resulting \emph{vulnerability window} falls from $25\%$ to
$7\%$ with model size, which is what makes small judges fragile. The churn stays within noise under
benign best-of-16 and becomes clear under adversarial pressure for some quantizers ($146$:$54$
harmful:helpful flips on one family, reversing sign on another). Downstream it changes what ships:
across six (model, quantizer) pairs from two families, $14.6$--$65.7\%$ of problems get a different
best-of-$64$ selection while only $2.6$--$9.6\%$ get a different final answer, with selections changing
$5$--$8\times$ more often than answers in every pair, and the pair the benchmark calls identical
($0.03$ points, $p=1.00$) still changes $34.5\%$ of them. Benchmark \emph{churn} ranks that
downstream change ($\rho=0.89$); benchmark \emph{accuracy} does not order it reliably. Held-out
audits then locate the useful diagnostic in the margin rather than the Gaussian: every
margin-based predictor orders flips identically (AUC $0.927$), and on the level a directly measured
churn rate wins in-distribution. As a routing signal it is matched by plain top-$k$. The
contribution is the characterization, not the model.
\end{abstract}

\section{Introduction}

Reward models (RMs) and LLM judges are infrastructure: they give the training signal in RLHF, pick
the winner in best-of-$N$ sampling, filter training data, and stand in for human raters. In all of
these roles the output is not a reported number but a \emph{decision}. Because they score many
candidates, they are among the first things practitioners quantize, and the usual justification for
\texttt{load\_in\_4bit=True} is that the benchmark number barely moves. We ask what that one-line
change actually does, and find the benchmark number is the wrong thing to watch.

\textbf{Aggregate accuracy is nearly blind to what quantization does to a reward model.} A
Skywork-Reward-V2 Llama-3.2-1B judge quantized to \nf{} scores $88.38\%$ on RewardBench against
$88.41\%$ in BF16. Underneath, $4.0\%$ of its pairwise decisions have changed; the errors it fixes
and introduces almost exactly cancel (60 vs.\ 59, $p=1.00$). Accuracy is a \emph{net} quantity,
churn a \emph{gross} one, and they come apart whenever induced errors are near-symmetric, which is the
default expectation absent any reason for quantization to prefer the correct answer.

We then explain the effect. Modeling quantization as additive noise on the score difference gives a
one-parameter prediction: a comparison with full-precision margin $m$ flips with probability
$\Phi(-|m|/\sig)$, where $\sig$ is the standard deviation of the induced perturbation. The empirical
curves lie close to $\Phi(-z)$ (Fig.~\ref{fig:law}a), $89$--$100\%$ of flips occur at $z<2$, and
with $\sig$ estimated on held-out pairs the prediction exceeds the observed rate in \emph{all 30}
scalar settings, an empirical envelope rather than just a fit.

The model explains the scaling we observe. What governs fragility is the scale-free ratio of
quantization noise to the model's own margin spread, $\sig/\mathrm{std}(m)$, which falls
$3\times$ from 0.6B to 8B within a single architecture; operationally, the share of decisions
inside the vulnerable band, the \emph{vulnerability window}, falls from $24.9\%$ to $7.0\%$.
Small reward models are fragile because they make more of their decisions within reach of their
own noise floor.

Whether this matters depends on what is downstream. Under benign best-of-16 the harm is within
noise; its point estimate increases with $N$, though our intervals do not establish growth. Under
\emph{adversarial} pressure it is clear: on 1{,}260 problems carrying verified-wrong but persuasive
candidates, \rtn{} makes every sub-8B judge easier to fool (146 harmful vs.\ 54 helpful flips;
problem-level 95\% CI on the difference $[63, 122]$) while an 8B control stays null. The effect is
quantizer-specific: it replicates directionally for \rtn{} on a second family and
\emph{reverses} for \nf{}.
\S\ref{sec:harm} shows the asymmetry is a base-rate effect (per comparison, originally-incorrect
decisions flip \emph{more} often), and that margin contraction, though real, does not improve
held-out flip prediction.

The downstream consequence is that the system returns a different output. Two versions of one
reward model separated by $0.03$ RewardBench points ($p=1.00$) pick different best-of-$64$
candidates on half of all problems, mostly answer-equivalent ones, with the extracted answer
changing on $4.5\%$ and the accuracy difference unresolved (\S\ref{sec:ship}). We then ask what the
noise model is for, and answer with held-out audits rather than assertion. \emph{Selective
escalation} scores all candidates with the quantized proposer and re-scores only comparisons with
$z<\tau$ at full precision: with a 0.6B \rtn{} proposer and an 8B BF16 verifier at $N=32$,
escalating $25.3\%$ of comparisons recovers $84\%$ of the accuracy gap at an estimated $52\%$ lower
scoring cost, and escalating to the proposer's \emph{own} BF16 recovers $80\%$ of the quantization
damage. But a plain top-$k$ rule matches or beats it at every budget, and a BF16 proposer recovers
most of the gap with no quantization to fix, so escalation is a demonstration that the window is
where the damage lives, not a method.

\paragraph{Contributions.}
\emph{(1) Churn that accuracy does not show, and where it sits.} We measure decision churn across
eight reward models, three generative judges and four quantizers, and characterize it: flips
concentrate at low normalized margin in every setting, a Gaussian approximation
$P(\text{flip})=\Phi(-|m|/\sig)$ over-predicts the \emph{aggregate} rate in all 30 scalar settings
while falling short for judges, and fragility tracks relative noise $\sig/\mathrm{std}(m)$ and
window occupancy (\S\ref{sec:churn}--\S\ref{sec:law}).
\emph{(2) When churn becomes consequence.} Benign best-of-16 harm is within noise, with point
estimates that increase with $N$ but intervals that do not establish growth; adversarial pressure
makes it clear for some quantizers and reverses it for others, and the harm asymmetry is a base-rate
effect rather than a tilt toward wrong answers. Downstream, across six (model, quantizer) pairs from
two families, selections change $5$--$8\times$ more often than final answers, and benchmark churn
ranks that change where benchmark accuracy does not (\S\ref{sec:harm}--\S\ref{sec:ship}).
\emph{(3) What to measure instead.} Held-out audits separate ordering from level: every
margin-based predictor orders flips identically, so the Gaussian contributes nothing there, while
on the level a directly measured churn rate beats it in-distribution and a logistic in $|m|$ matches
it across distributions. As a routing signal it is matched by plain top-$k$. We also report what did
not work and give a claim-to-evidence table (\S\ref{sec:audit}--\S\ref{sec:cascade},
Appendices~\ref{app:negative} and~\ref{app:checklist}).

\section{Related Work}

\paragraph{Post-training quantization.}
Post-training quantization of LLMs spans 8-bit mixed-precision and weight-activation schemes
\citep{dettmers2022llmint8,xiao2023smoothquant} and the weight-only 4-bit methods that are now
standard practice \citep{frantar2023gptq,lin2024awq,dettmers2023qlora,shao2024omniquant}, and is
almost universally evaluated on perplexity or downstream accuracy of \emph{generative} models. We study models whose output is a scalar used to
make a comparison, where what matters is the stability of an \texttt{argmax}, not the fidelity of a
distribution.

\paragraph{Reward models and overoptimization.}
Reward models underpin RLHF \citep{christiano2017preferences,stiennon2020summarize,ouyang2022instructgpt}
and are increasingly used directly as judges \citep{zheng2023judging}. RewardBench
\citep{lambert2024rewardbench} is the standard pairwise benchmark and reports aggregate accuracy,
exactly the metric we show to be insensitive. \citet{gao2023scaling} established that optimizing
against an imperfect proxy degrades true reward, with ensembling studied as a mitigation
\citep{coste2024ensembles,eisenstein2023helping}. Our contribution is orthogonal: quantization is a
distinct, controllable, previously unmeasured source of proxy error, and we characterize exactly
which decisions it corrupts. That optimizers exploit reward misspecification is long-standing
\citep{amodei2016concrete,skalse2022defining}; LLMBar \citep{zeng2024llmbar} is the RewardBench
subset on which we observe the highest churn.

\paragraph{Decision-level effects of quantization.}
Three lines of work are close to ours. QDPO \citep{lee2024qdpo} studies token flipping in
quantized \emph{generative} LLMs and repairs it by preference optimization; we study decisions,
not tokens, and do not fine-tune. \citet{illusion2026equivalency} report behavioral divergence
under quantization despite preserved accuracy via a decision-agreement metric; our churn
measurement is that observation specialized to preference decisions, and our contribution beyond
it is the margin model that predicts \emph{which} decisions change. Most closely,
\citet{vecsearch2026quant}, contemporaneously and for vector search, decompose comparison flips
into near-zero-margin mass plus a calibrated residual tail, treat residual dependence, and give
held-out selective-risk bounds, a parallel framework in another domain. We claim no priority on
margin-based analysis of quantized comparisons; what is specific here is its characterization on
reward models and judges, the quantizer-dependent contraction, and that routing on it does not
beat top-$k$.

\paragraph{Cascades.}
Routing easy inputs to a cheap model is well established, from speculative decoding
\citep{leviathan2023speculative,chen2023accelerating} to cost-aware cascades
\citep{chen2023frugalgpt} and learned routers trained on preference data \citep{ong2025routellm}.
Ours differs in where the routing signal comes from: not a learned router or heuristic confidence,
but a calibrated noise model giving an empirical envelope on residual disagreement with the
proposer's own full-precision counterpart, which, as \S\ref{sec:cascade} shows, is all it adds
over a plain top-$k$ rule.

\section{Setup}
\label{sec:setup}

\paragraph{Models and quantizers.}
We use the Skywork-Reward-V2 family, which spans a wide size range within one training recipe:
Qwen3-based models at 0.6B, 1.7B, 4B, 8B and Llama-3.2-based models at 1B and 3B, all
sequence-classification RMs emitting a single scalar. BF16 is ground truth throughout: we measure
what quantization \emph{changes}, not what is correct. We evaluate four weight-only 4-bit schemes:
\nf{} and \textsc{FP4} (bitsandbytes, double quantization, block 64), where \textsc{FP4} is the
documented default of \texttt{load\_in\_4bit=True}; \nf{} requires explicit configuration; \rtn{},
symmetric round-to-nearest at group 128; and \gpt{} at group 128, calibrated on 512 UltraFeedback
chat pairs and 1{,}200 GSM8K \emph{training}-split pairs, disjoint from every evaluation set. All
quantize linear layers only. We additionally score three \emph{generative} pairwise judges
(\S\ref{sec:law}), whose margin is the difference of the ``A'' and ``B'' verdict logits averaged
over both response orderings (Appendix~\ref{app:judge}).

\paragraph{Evaluations and metrics.}
\emph{(i)} RewardBench \citep{lambert2024rewardbench}, filtered split, 2{,}985 pairs over 23
subsets. \emph{(ii)} Adversarial best-of-4: 1{,}260 problems (329 MATH-500, 931 GSM8K), each with
one mechanically verified-correct solution and three verified-\emph{wrong} but persuasive ones
written by a Qwen2.5-7B adversary; since correctness is checked by symbolic answer matching,
``fooled'' is an objective event. \emph{(iii)} Large-$N$ selection: 1{,}000 problems (all 500 of
MATH-500 plus 500 from GSM8K) with 64 sampled candidates each (Appendix~\ref{app:downstream}). For
a pair with chosen response $a$ and rejected response $b$, the \emph{margin} is
$m=s_a-s_b$, so $m>0$ is a correct decision; a \emph{flip} is a comparison whose
$\operatorname{sign}(m)$ differs between BF16 and the quantized model. We distinguish the
\emph{latent} normalized margin $z=|m_{\text{BF16}}|/\sig$, used to state and validate the model, from
the \emph{observed} statistic $\hat z=|m_{\text{quant}}|/\sig$, which is all an inference-time
system can compute (\S\ref{sec:cascade}). We report
flip rate with a bootstrap $95\%$ CI and split flips into \emph{harmful} (BF16 right, quantized
wrong) and \emph{helpful}, testing asymmetry with an exact two-sided McNemar test. The \emph{noise
scale} is $\sig=\operatorname{std}(m_{\text{quant}}-m_{\text{BF16}})$; for the predictive test in
\S\ref{sec:law} it is estimated on even-indexed RewardBench pairs and evaluated on odd-indexed
ones, so $\sig$ is never fit to the comparisons it predicts. Because models share problems and each
trap contributes three comparisons, pooled harm:help results and all best-of-$N$ differences use a
\emph{problem-level clustered bootstrap}: 2{,}000 resamples of problems, with percentile intervals
(and two-sided bootstrap $p$) over new problems, not over comparisons or candidate subsets. Of the
66 McNemar tests reported, 33 of the 41 reaching $p<0.05$ survive Benjamini--Hochberg at $q<0.05$;
we mark the eight that do not with $\dagger$.
All runs use a single H100 NVL (94GB); every number and figure is regenerated from the scored
outputs by \texttt{scripts/make\_paper\_assets.py}.

\begin{table}[t]
\centering\small
\caption{\textbf{Decision churn is invisible to accuracy.} 4-bit quantization changes a large
fraction of RewardBench decisions while aggregate accuracy moves little. The Llama-3.2-1B/\nf{} row
is the extreme case: $3.99\%$ of decisions flip and accuracy moves $0.03$ points, with harmful and
helpful flips almost perfectly balanced. Harm:help is (BF16 right, quantized wrong):(the reverse);
$p$ is an exact two-sided McNemar test. $n=2{,}985$ pairs.}
\label{tab:churn}
\setlength{\tabcolsep}{2.6pt}
\begin{tabular}{llccccccc}
\toprule
Model & BF16 acc. & Quantizer & Acc. & $\Delta$acc. (pt) & Flip \% & 95\% CI & Harm:help & $p$ \\
\midrule
\multirow{4}{*}{Qwen3-0.6B} & 87.54 & NF4 & 85.46 & -2.08 & 6.83 & \footnotesize[5.93,7.74] & 133:71 & \textbf{$<$0.001} \\
 &  & FP4 & 82.04 & -5.49 & 12.26 & \footnotesize[11.09,13.40] & 265:101 & \textbf{$<$0.001} \\
 &  & INT4-RTN & 82.58 & -4.96 & 12.33 & \footnotesize[11.22,13.54] & 258:110 & \textbf{$<$0.001} \\
 &  & GPTQ & 84.76 & -2.78 & 7.67 & \footnotesize[6.77,8.61] & 156:73 & \textbf{$<$0.001} \\
\midrule
\multirow{4}{*}{Llama-3.2-1B} & 88.41 & NF4 & 88.38 & -0.03 & 3.99 & \footnotesize[3.28,4.66] & 60:59 & 1.00 \\
 &  & FP4 & 86.26 & -2.14 & 6.97 & \footnotesize[6.03,7.91] & 136:72 & \textbf{$<$0.001} \\
 &  & INT4-RTN & 86.87 & -1.54 & 7.04 & \footnotesize[6.20,8.01] & 128:82 & \textbf{0.002} \\
 &  & GPTQ & 84.19 & -4.22 & 9.18 & \footnotesize[8.17,10.22] & 200:74 & \textbf{$<$0.001} \\
\midrule
\multirow{4}{*}{Qwen3-1.7B} & 92.06 & NF4 & 91.09 & -0.97 & 3.58 & \footnotesize[2.95,4.32] & 68:39 & \textbf{0.007} \\
 &  & FP4 & 89.45 & -2.61 & 5.43 & \footnotesize[4.62,6.23] & 120:42 & \textbf{$<$0.001} \\
 &  & INT4-RTN & 88.74 & -3.32 & 6.20 & \footnotesize[5.33,7.04] & 142:43 & \textbf{$<$0.001} \\
 &  & GPTQ & 90.59 & -1.47 & 4.49 & \footnotesize[3.82,5.26] & 89:45 & \textbf{$<$0.001} \\
\midrule
\multirow{4}{*}{Llama-3.2-3B} & 93.20 & NF4 & 92.29 & -0.90 & 2.38 & \footnotesize[1.84,2.91] & 49:22 & \textbf{0.002} \\
 &  & FP4 & 91.66 & -1.54 & 3.55 & \footnotesize[2.91,4.22] & 76:30 & \textbf{$<$0.001} \\
 &  & INT4-RTN & 92.86 & -0.34 & 3.02 & \footnotesize[2.45,3.62] & 50:40 & 0.34 \\
 &  & GPTQ & 91.89 & -1.31 & 2.85 & \footnotesize[2.28,3.45] & 62:23 & \textbf{$<$0.001} \\
\midrule
\multirow{4}{*}{Qwen3-4B} & 94.91 & NF4 & 94.47 & -0.44 & 1.24 & \footnotesize[0.84,1.64] & 25:12 & \textbf{0.047}$^{\dagger}$ \\
 &  & FP4 & 94.34 & -0.57 & 1.91 & \footnotesize[1.41,2.45] & 37:20 & \textbf{0.033}$^{\dagger}$ \\
 &  & INT4-RTN & 94.30 & -0.60 & 2.28 & \footnotesize[1.74,2.85] & 43:25 & \textbf{0.038}$^{\dagger}$ \\
 &  & GPTQ & 94.34 & -0.57 & 1.78 & \footnotesize[1.34,2.28] & 35:18 & \textbf{0.027}$^{\dagger}$ \\
\midrule
\multirow{4}{*}{Qwen3-8B} & 95.08 & NF4 & 94.87 & -0.20 & 1.21 & \footnotesize[0.84,1.61] & 21:15 & 0.41 \\
 &  & FP4 & 94.94 & -0.13 & 1.41 & \footnotesize[1.01,1.84] & 23:19 & 0.64 \\
 &  & INT4-RTN & 94.34 & -0.74 & 1.61 & \footnotesize[1.17,2.08] & 35:13 & \textbf{0.002} \\
 &  & GPTQ & 94.97 & -0.10 & 1.11 & \footnotesize[0.74,1.54] & 18:15 & 0.73 \\
\bottomrule
\end{tabular}

\end{table}

\section{Decision Churn Is Invisible to Accuracy}
\label{sec:churn}

Table~\ref{tab:churn} reports, for every model and quantizer, what the benchmark says and what
actually changed. The two disagree badly.

The clearest case is Llama-3.2-1B under \nf{}: accuracy moves $-0.03$ points and McNemar returns
$p=1.00$ (60 decisions worse, 59 better), yet $3.99\%$ of decisions, 119 of 2{,}985, are
different. Nor is it isolated. Qwen3-8B under \nf{} flips $1.21\%$ for $-0.20$ points ($p=0.41$),
and Llama-3.2-3B under \rtn{} flips $3.02\%$ for $-0.34$ points ($p=0.34$). In each, accuracy is
statistically flat while a few percent of the decision function has been rewritten.

Churn is not damage, and we do not claim it is: \S\ref{sec:harm} shows the flips are near
symmetric under benign selection at small $N$ and that some quantizers are protective. It means the deployed evaluator is not
the one that was benchmarked, and the benchmark cannot detect the substitution.

\begin{figure}[t]
\centering
\includegraphics[width=0.85\linewidth]{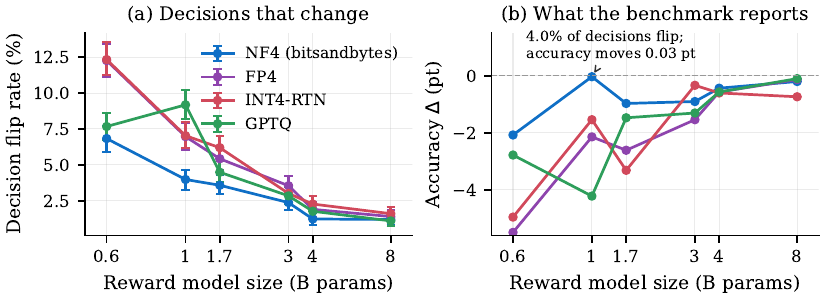}
\caption{\textbf{(a)} Decision flip rate falls with reward model size (bootstrap 95\% CIs);
monotone for \nf{}, FP4 and \rtn{}, with one \gpt{} exception at Llama-3.2-1B. \textbf{(b)} The accuracy delta that a practitioner would actually
look at shows no such structure and is near zero in several of the highest-churn settings.}
\label{fig:churn}
\end{figure}

Fig.~\ref{fig:churn} shows the pattern across sizes. Flip rates fall with model size for every
quantizer, monotonically for \nf{}, FP4 and \rtn{}, from $6.83\%$ (\nf{}) and $12.33\%$
(\rtn{}) at 0.6B to $1.21\%$ and $1.61\%$ at 8B, with one \gpt{} exception that rises from
$7.67\%$ at 0.6B to $9.18\%$ at Llama-3.2-1B (\S\ref{sec:harm}), while accuracy deltas wander between $-5.49$ and $-0.03$ points with no clean
structure. The churn signal is orderly; the accuracy signal is not.

Churn is also not uniform over the benchmark: it concentrates on adversarial subsets, reaching
$12$--$13\%$ on \texttt{llmbar-adver-*} and up to $8.5\%$ on code at 1B against low single digits on
easy chat, since the comparisons quantization corrupts are the ones the benchmark includes because they
are hard. And \textbf{quantizer choice matters more than model choice}: \rtn{} flips
$1.3$--$1.8\times$ as many decisions as \nf{} on the same model.

\section{An Additive-Noise Model of Quantization Flips}
\label{sec:law}

\subsection{Derivation and validation}

\begin{figure}[t]
\centering
\includegraphics[width=0.85\linewidth]{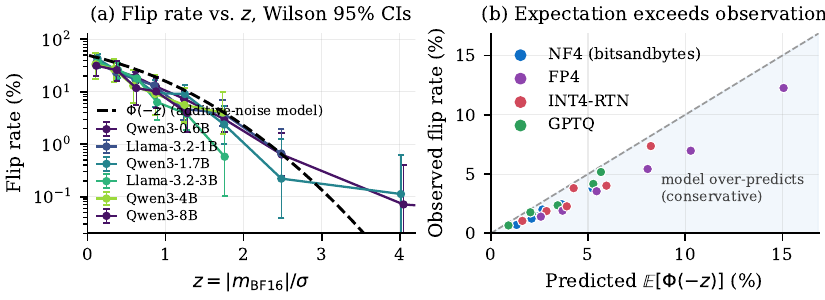}
\caption{\textbf{(a)} Flip rate versus normalized margin $z=|m_{\mathrm{BF16}}|/\sigma$ for all
six Skywork models under \nf{}. After rescaling by each model's own $\sigma$, the curves lie
close to the additive-noise prediction $\Phi(-z)$ (dashed). This is one family under one quantizer, not
a claim of exact collapse. \textbf{(b)} Predicted versus observed
aggregate flip rate for all 30 model\,$\times$\,quantizer settings. Every point lies below the
diagonal: the Gaussian expectation over-predicts, so the envelope is conservative for these scalar
models.}
\label{fig:law}
\end{figure}

Let $m$ be the BF16 margin and $\tilde m = m + \varepsilon$ the margin after quantization. Modeling
$\varepsilon$ as zero-mean Gaussian with scale $\sig$, independent of $m$, the comparison flips
exactly when $\varepsilon$ crosses $-m$:
\begin{equation}
P(\text{flip} \mid m) \;=\; \Phi\!\left(-\frac{|m|}{\sig}\right) \;=\; \Phi(-z).
\label{eq:law}
\end{equation}
Eq.~\eqref{eq:law} has one parameter, $\sig$, measured directly from the two score files and never
fit to flip data. It predicts that after rescaling each model's margins by its own $\sig$, every
model and quantizer lies on the \emph{same} curve. Fig.~\ref{fig:law}a confirms this for all six
models under \nf{}, over 6{,}765 pooled comparisons per model (RewardBench pairs plus every
correct-vs-trap comparison from the adversarial pools, so both benign and adversarial margin regimes
are represented): the rescaled curves lie close to $\Phi(-z)$ across three orders of magnitude.

Table~\ref{tab:law} (Appendix~\ref{app:law}) gives the quantitative version. Across all 30 model\,$\times$\,quantizer
settings, $88.9$--$100\%$ of every flip occurs at $z<2$, the \textbf{vulnerability window},
comparisons whose full-precision margin is within two units of the model's own quantization-noise
scale. At inference time only $\hat z=|m_{\text{quant}}|/\sig$ is observable, and
\S\ref{sec:cascade} uses it as a proxy. Under its assumptions Eq.~\eqref{eq:law} is an equality, not a bound;
what we observe empirically is that its expectation $\mathbb{E}[\Phi(-z)]$ \emph{exceeds} the
measured flip rate in every one of the 30 scalar settings (Fig.~\ref{fig:law}b), typically by
$20$--$45\%$ relative, and this holds with $\sig$ estimated on held-out pairs, $30/30$
(Appendix~\ref{app:law}), while it \emph{falls short} for generative judges (below). The Gaussian
model is thus an idealization whose expectation happens to be a conservative envelope for these
scalar architectures, on an aggregate rate rather than a per-comparison conditional. We leave the
slack unexplained ($\sig$ already absorbs the margin-error scale, so candidate-error correlation
does not account for it) and use the envelope as a heuristic, not a theorem.

\paragraph{The model is not a property of one model family.}
Everything above uses Skywork-Reward-V2, so we replicated on two GRM reward models
\citep{yang2024grm}, trained with a different recipe and objective
(Appendix~\ref{app:repl}). Every prediction transfers: $95.7$--$97.7\%$ of flips fall at $z<2$,
$\Phi(-z)$ over-predicts in all four, and the churn is again invisible: GRM-Llama-3.2-3B under
\nf{} flips $2.68\%$ of its decisions for $-0.34$ points ($p=0.31$).

\paragraph{Generative judges: the phenomenon extends, the bound does not.}
A \emph{generative} pairwise judge yields a margin as the difference of its ``A'' and ``B'' verdict
logits, so the same statistics apply. On three instruct models (Appendix~\ref{app:judge}) churn
replicates and occupancy predicts its size ($33\%$, $55\%$, $90\%$ occupancy for Qwen2.5-7B,
Mistral-7B, Qwen2.5-3B, with flip rates $6.4\%$, $8.0\%$, $37.6\%$), and the strongest judge
gives the purest instance of the central effect: \rtn{} rewrites $6.37\%$ of Qwen2.5-7B's
decisions while accuracy moves $+0.07$ points ($94$:$96$, $p=0.94$). The \emph{bound} does not
transfer: $\Phi(-z)$ under-predicts in two of six judge settings, badly for Qwen2.5-3B/\rtn{}
($23.6\%$ predicted, $37.6\%$ observed), because judges flip at $z=3$--$5$ where Gaussian noise
predicts almost none. We therefore scope the envelope of \S\ref{sec:cascade} to scalar reward
models, where it held in all 30 settings.

\subsection{Why small models are fragile}

Eq.~\eqref{eq:law} separates damage into the noise the quantizer injects and how exposed the model's
decisions are to it. Fig.~\ref{fig:mechanism} shows both fall with scale once measured in
scale-free units.

Raw $\sig$ is confounded by each architecture's output scale, so pooling families hides the trend;
within a family it is clean. Under \nf{} the scale-free ratio $\sig/\mathrm{std}(m_{\text{BF16}})$
falls monotonically in Qwen3 from $0.34$ (0.6B) to $0.10$ (8B), and its operational form, window
occupancy $P(z<2)$, falls the same way, $24.9\%\to7.0\%$ (Fig.~\ref{fig:mechanism},
Appendix~\ref{app:law}). Larger models separate their preferences by more multiples of their own
noise, and occupancy predicts damage better than raw $\sig$ ($r=0.96$ vs.\ $0.78$).

The paper's mechanistic claim in one sentence: \textbf{quantization is additive decision noise,
and a model's fragility is set by the size of that noise relative to its own margin spread,
which falls with scale, so small reward models make a larger fraction of their decisions within
reach of their noise floor.}

\section{When Churn Becomes Harm}
\label{sec:harm}

Churn is instability, not damage. A flipped decision on a pair the model was guessing at costs
nothing. Whether it matters depends on what consumes the decisions.

\paragraph{Under benign selection at small $N$, barely.}
On 2{,}319 benign math problems (MATH-500 over two seeds plus GSM8K, 16 candidates each) with
\nf{}, no individual model shows significant harm (31:25, 35:22, 19:13, 19:20 from 0.6B to 3B, all
$p\ge0.11$) and pooled the split is 104:80 ($p=0.09$), which is what Eq.~\eqref{eq:law} predicts for
$N=16$, where margins against sampled alternatives are wide. \textbf{At best-of-16 quantization
harm is within noise; regret point estimates increase with $N$, but the current uncertainty does
not establish growth}; a stronger claim would need a paired test across $N$, which we do not run.

\paragraph{Under adversarial pressure, the effect is real but quantizer-specific.}
Table~\ref{tab:adv} and Fig.~\ref{fig:harm} (Appendix~\ref{app:harmfig}) report adversarial best-of-4, where every wrong
candidate is verifiably wrong but written to be persuasive. \rtn{} increases the fooling rate on
\emph{both} families: pooled over the four sub-8B Skywork models it gives 146 harmful versus 54
helpful flips (problem-level clustered bootstrap, 2{,}000 resamples of problems: 95\% CI on
harmful$-$helpful $[63, 122]$; the unclustered McNemar $p<10^{-10}$ over-states
this since the four models share problems), and over the two GRM models 51:31 (CI
$[2, 38]$, bootstrap $p=0.031$; the McNemar $p=0.035$ does not survive
Benjamini--Hochberg across 66 tests), so we claim replication of the \emph{direction} on GRM,
marginally. At 0.6B
the fooling rate rises from $5.87\%$ in BF16 to $10.48\%$, a $1.8\times$ increase.

\nf{} does \emph{not} transfer: it pools to 73:35 in the harmful direction on Skywork
($p<10^{-3}$) but 13:41 in the \emph{opposite} direction on GRM ($p<10^{-3}$), having made those
models measurably \emph{harder} to fool: quantization does not uniformly make evaluators easier to
exploit.

\paragraph{Why: exposure, base rates, and a one-sided residual.}
Eq.~\eqref{eq:law} models quantization as zero-mean noise, which predicts how \emph{many} decisions
flip but not \emph{which way}. Three held-out facts constrain the explanation
(Appendix~\ref{app:bias}, Tables~\ref{tab:contract}--\ref{tab:threeparam}).
\emph{First, contraction is an observed component but not a better predictor.} \rtn{} shrinks
margins (slope $-0.41$ at 0.6B) with near-zero intercept, so its negative mean shift is mostly
scale; yet fitting $\tilde m = c\,m + b + \eta$ on even-indexed comparisons and predicting
odd-indexed ones, the three-parameter model wins only $17/30$ settings and is \emph{worse} for
\rtn{} in five of eight. \emph{Second, the count asymmetry is base rates, and the per-comparison
asymmetry runs the other way}: $68$--$84\%$ of in-window comparisons were originally correct, so
more correct decisions are available to flip, yet originally-\emph{incorrect} ones flip
$1.5$--$3\times$ more often ($30.7\%$ vs.\ $16.1\%$ at 0.6B/\rtn{}). \emph{Third, the envelope's
conservatism is one-sided}: held-out, it over-predicts harmful flips by $1.1$--$2.4\times$ while
the helpful side lies within $\pm20\%$ in $19$ of $30$ settings. The 8B control is unaffected by
either quantizer (\rtn{} $10$:$7$; \nf{} $6$:$5$).

\paragraph{Direction is model- and distribution-specific and does not transfer.}
The mean shift's sign agrees with the direction of harm in 19 of 23 settings ($r=-0.55$), but both
come from the same margins; out of sample the RewardBench shift predicts trap-harm direction at
only $15/19$ ($r=-0.39$, $p=0.10$) and is negative in every setting, so the sign test has no power.
What is robust is the same-data contrast: under the same quantizer and traps, Qwen2.5-7B becomes
easier to fool ($73$:$35$; clustered bootstrap CI $[18, 58]$) while Mistral-7B becomes \emph{harder}
($61$:$125$, $[-90, -39]$; Appendix~\ref{app:judge}). The practitioner-facing claim is narrow: \textbf{coarse
round-to-nearest raises exposure most, and whether flips net out as harm must be measured where you
deploy} (Appendix~\ref{app:bias}).

\paragraph{Search amplifies the damage, with wide uncertainty.}
Point estimates of best-of-$N$ regret grow with $N$, but with problem-level intervals only the
0.6B \rtn{} regret at $N=4$ and $N=16$ excludes zero ($0.72$, $[0.06, 1.43]$ at $N=16$); at $N=64$
it is $1.20$ $[-0.31, 2.79]$ and GRM never excludes zero (Appendix~\ref{app:harmfig}). We report
the point-estimate trend and do not claim growth is established.

\paragraph{Better quantizers help, but not reliably.}
\gpt{} recovers most of \rtn{}'s trap damage (at 0.6B, fooling $10.48\%\to6.98\%$ against a
BF16 floor of $5.87\%$; Table~\ref{tab:adv}). But not \emph{reliably}: on RewardBench,
Llama-3.2-1B under \gpt{} is \emph{worse} than under \rtn{} ($9.18\%$ flips vs.\ $7.04\%$,
$-4.22$ vs.\ $-1.54$ points; Table~\ref{tab:churn}) despite reconstructing $3\times$ better
(Appendix~\ref{app:gptqval}). Calibration-based quantizers minimize reconstruction error, an
objective unaware of decision margins.

\subsection{What you ship: which output the selector returns}
\label{sec:ship}

The results so far concern comparisons; a practitioner asks whether the returned \emph{output}
changes. We test the labeller for which the benchmark is most confidently blind (Llama-3.2-1B
under \nf{}, $-0.03$ points at $p=1.00$, Table~\ref{tab:churn}), letting each version select a
best-of-$N$ answer from the same 64 candidates on the same 1{,}000 problems. Mechanically verified
correctness is the objective endpoint; the arbiter, Skywork-8B BF16, is a larger evaluator from the
same family rather than independent human judgement, so its preference is corroboration only. Both
selectors see identical draws and share one deterministic tie-breaking rule, which matters because
BF16 score precision leaves $32$--$39$ distinct values per problem and ties the top score on
$12$--$15\%$. Intervals resample whole problems, stratified by dataset
(Appendix~\ref{app:downstream}).

\textbf{The selector returns a different output very often, and an answer-equivalent one most of
the time} (Table~\ref{tab:ship}). At $N=64$ the two versions select different candidates on
$50.3\%$ of problems, but $90.7\%$ of those switches return the \emph{same} final answer, and
$31.4\%$ of them occur where a maximum was tied, so the rate of disagreement about a \emph{unique}
best candidate is $34.5\%$. Extracted answers differ on $4.5\%$ of problems. Correctness changes on
only $4.0\%$ of switches ($2.8\%$ toward BF16, $1.2\%$ toward \nf{}); accuracy moves $+0.80$ points
at $N=64$ ($[-0.10,1.70]$), the arbiter prefers BF16 on $49.7\%$ of disagreements, and answer
divergence is dataset-dependent ($6.8\%$ MATH-500 vs.\ $2.2\%$ GSM8K at $N=32$).

\paragraph{This is not one pair.} The identical protocol (same pools, draws and tie rule, each
quantized model against \emph{its own} BF16) run over six (model, quantizer) pairs from two
families and three quantizers (Table~\ref{tab:matrix}) gives $14.6$--$65.7\%$ different
unique-maximum selections at $N=64$ against only $2.6$--$9.6\%$ different answers: selections change
$5.3$--$7.7\times$ more often than answers in \emph{every} pair. \textbf{Large identity churn with
an order of magnitude less answer churn is the general behaviour, not a property of one pair.} What
ranks that change is benchmark \emph{churn}, not benchmark \emph{accuracy} ($\rho=0.89$, $p=0.019$
against $\rho=0.77$, $p=0.07$). Six pairs is too few to lean on, but the ordering fails
concretely: Llama-3.2-1B/\nf{} loses $0.03$ RewardBench points and changes $34.5\%$ of selections
while GRM-gemma2-2B/\rtn{} loses $28\times$ more and changes $22.1\%$.

\textbf{Quantizing a reward model substantially changes \emph{which} output a best-of-$N$
system returns while usually leaving the answer unchanged, and our data do not
resolve whether the selectors differ in quality.} Across all six pairs and four values of $N$, only
$4$ of $24$ accuracy intervals exclude zero, three of them the same pair, and all four are
small ($+0.33$ to $+1.03$ points, every one favouring BF16). That is an unresolved comparison, not
equivalence, which would need an application-justified tolerance and a paired test. What is
established is that a $0.03$-point benchmark difference coexists with a third of the returned
outputs changing identity at a unique maximum, or half of them counting tie-breaks.

\section{What Is the Model For?}
\label{sec:audit}

Such a model has two plausible uses: \emph{auditing} (predicting instability on unlabelled
problems) and \emph{routing}. Both are largely negative, with one dataset-specific exception.

\subsection{Audit: when calibration and deployment match, measure rather than model}

Split \emph{problems} into 200 or 500 calibration and a held-out remainder (a trap problem's three
comparisons stay together); estimate $\sig$ on calibration, freeze it, predict held-out flips as
$\Phi(-|m_{\text{BF16}}|/\sig)$. Baselines on the same data: the calibration churn rate, or
$P(\text{flip})=\text{sigmoid}(a+b|m|)$; three seeds, none dropped (Appendix~\ref{app:audit}).

Within one (model, quantizer) all three margin-based predictors are monotone transforms of $|m|$,
so they order comparisons identically (AUC $0.927$ on RewardBench and $0.981$ on the traps,
agreeing to machine precision), while the constant rate orders nothing (AUC $0.500$). The margin
is a strong indicator of \emph{which} comparisons flip, and neither $\sig$ nor the Gaussian form
contributes to that ordering; these predictors can differ only in the probability \emph{level} they
attach. On the level the $\sig$ model loses: $2.35$ points against $0.99$ for simply reporting the
calibration churn rate, $79$ of $90$ settings ($85$ of $90$ at 500), with a logistic in $|m|$
better still ($0.96$). Brier mixes level with discrimination, so the $\sig$ model's $90/90$
advantage there over the constant rate reflects that baseline's complete lack of discrimination,
and its residual gap to the logistic ($0.036$ vs.\ $0.035$) is calibration alone. The ordering
reverses on the traps at 200 calibration problems ($0.21$ vs.\ $0.47$) but the gap closes by 500
($0.23$ vs.\ $0.27$; logistic $0.22$). \textbf{To learn how much your evaluator churns on data like
your calibration set, measure it; to learn which comparisons churn, use the margin, to which the Gaussian
adds nothing.}

Transfer is where the Gaussian does best, and a churn rate structurally cannot transfer at all: fit
on RewardBench and applied to the traps the $\sig$ model errs by $0.59$ points against $3.40$;
reverse, $1.10$ against $3.40$. A paired bootstrap over source and target problems, refitting on
every draw, puts that advantage below zero in $16/19$ and $13/19$ settings. Refitting on 500 source
problems rather than the whole source reproduces the $\sig$ model's aggregate errors ($0.59$ and
$1.07$ against $0.59$ and $1.10$), so the advantage is not a calibration-budget artifact, though
those budget-matched rows are point estimates and carry no paired intervals. Against the logistic
the same paired test is a wash ($3/19$ and $2/19$ wins, $8/19$ and $0/19$ losses):
margin-conditioning is what transfers, not the Gaussian form. These are two directions between two
datasets, not independent replications.

\subsection{Routing: the window localizes the damage, but top-$k$ selects as well}
\label{sec:cascade}

A quantized proposer scores all $N$; the leader and every $c$ with
$\hat z=|s_q(c^\star)-s_q(c)|/\sig<\tau$ are re-scored at full precision. At $N=32$ this recovers
$84\%$ of the 0.6B-to-8B gap at an estimated $52.5\%$ lower scoring cost ($R=5.03$ measured, a
scoring-cost estimate rather than latency; the GRM pair, at $R=1.00$, replicates the recovery but
not the saving). Paired intervals split that gap into $0.99$ of quantization damage ($[0.07,1.96]$)
and $0.74$ of capacity ($[-0.21,1.73]$), and escalating to the proposer's \emph{own} BF16 recovers
$0.79$ of the $0.99$ ($[0.12,1.52]$), so the window does hold the quantization error. But the
routing is not what selects: $z<\tau$ is exactly $|m|<\tau\sig$, and at the same budget top-$k$ is
$0.35$ points better ($[-0.71,-0.00]$, Appendix~\ref{app:fidelity}).

\section{Limitations and Conclusion}

Scope: weight-only 4-bit, math-only best-of-$N$, one generator and one arbiter, judges evidenced
on traps, two reward-model families (Appendix~\ref{app:negative}).

That quantized models disagree with their full-precision counterparts is established
\citep{illusion2026equivalency}; a deployment needs to know where it sits and which gate to put it
behind. \emph{(i)} Size the risk by measuring, not modelling:
$200$--$500$ problems predict held-out churn to about a point and the noise model, given the same
data, does worse (\S\ref{sec:audit}); concentrate that audit on low-margin comparisons and on the
smallest evaluator in the stack, where $95.7$--$97.7\%$ of flips sit. \emph{(ii)} Gate at the answer level: on these math tasks selection
agreement flags $50.3\%$ of problems, benchmark accuracy flags none, and the quantity between them,
answer-level disagreement, is $4.5\%$, with selections outrunning answers $5$--$8\times$ in
all six pairs (\S\ref{sec:ship}). Screen on benchmark churn, not accuracy: churn ranks the
downstream change. \emph{(iii)} Harm has a sign, and it belongs to the pair rather than to
4-bit: \nf{} reverses between families, so re-measure per family (\S\ref{sec:harm}).

\section*{Reproducibility Statement}
All scoring outputs, analysis scripts, and the code that regenerates every table and figure in this
paper from those outputs are released alongside this preprint. \S\ref{sec:setup} states the models,
quantizer configurations, datasets and statistical tests; the noise scale $\sig$ and the flip,
harm:help and McNemar statistics are all recomputed from the raw per-comparison scores by
\texttt{scripts/make\_paper\_assets.py}, so no reported number is transcribed by hand.

\section*{Use of AI Assistance}
We disclose that a large language model
(Claude, Anthropic) was used extensively in producing this work, under the direction and
supervision of the authors. Specifically, the model: drafted and revised all sections of the
manuscript; designed and implemented the analysis pipeline that recomputes every reported
statistic, table and figure from the raw per-comparison scores; implemented the generative-judge
scoring procedure (\S\ref{sec:setup}), the top-$k$ and random escalation baselines
(Appendix~\ref{app:fidelity}), and the activation-weighted validation of our GPTQ implementation;
proposed several of the experiments reported here, including the independent-family replication,
the expansion of the best-of-$N$ pool from 200 to 1{,}000 problems, and the out-of-sample test that
led us to retract an earlier claim that the systematic shift \emph{predicts} the direction of harm
(\S\ref{sec:harm}); and identified that claim as circular. No experimental data were generated by
the model: every score in this paper comes from running the named reward models and judges on the
named datasets on our hardware. The authors set the research direction, executed all experiments,
made every decision about framing and scope, and verified every number and claim; we take full
responsibility for the contents of the paper.

\bibliographystyle{iclr2027_conference}
\bibliography{refs}

\appendix
\begin{figure}[H]
\centering
\includegraphics[width=0.90\linewidth]{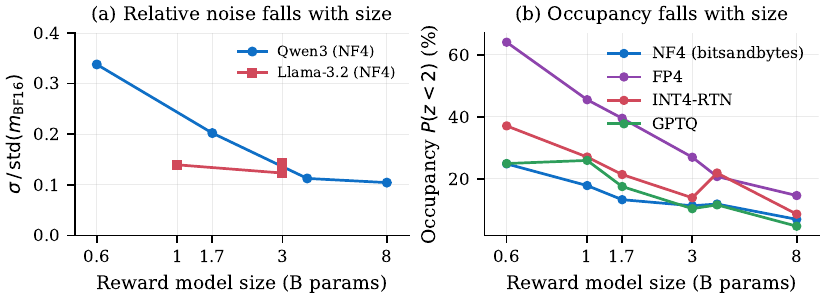}
\caption{\textbf{Relative noise and exposure both fall with size.} \textbf{(a)} The scale-free
noise ratio $\sig/\mathrm{std}(m_{\text{BF16}})$ within each family; raw $\sig$ is confounded by
architecture-specific logit scale. \textbf{(b)} Window
occupancy $P(z<2)$ falls steeply with size. Small models are relatively noisier \emph{and} more
exposed; the two are the same fact in different units.}
\label{fig:mechanism}
\end{figure}

\section{Full Decomposition Table}
\label{app:law}

Table~\ref{tab:law} gives the per-setting decomposition summarized in \S\ref{sec:law}.
Table~\ref{tab:heldout} gives the held-out check: $\sig$ estimated on even-indexed RewardBench
pairs only, $\mathbb{E}[\Phi(-z)]$ evaluated against the observed flip rate on odd-indexed pairs,
with a bootstrap interval on the difference. The expectation exceeds the observation in all 30
scalar settings; the interval excludes zero in every \nf{} and \rtn{} setting shown.

\begin{table}[H]
\centering\small
\setlength{\tabcolsep}{4pt}
\caption{\textbf{Held-out envelope.} $\sig$ from even-indexed RewardBench pairs; prediction and
observation on odd-indexed pairs; difference with a 2{,}000-draw bootstrap 95\% interval over
test pairs. (\nf{} and \rtn{} shown; FP4 and \gpt{} rows in \texttt{tables.json}.)}
\label{tab:heldout}
\begin{tabular}{lcccccc}
\toprule
Model / quant. & $n_{\text{cal}}$ & $n_{\text{test}}$ & $\sigma_{\text{cal}}$ & pred.\ \% & obs.\ \% & diff.\ [95\% CI] \\
\midrule
Qwen3-0.6B / nf4 & 1500 & 1485 & 1.58 & 11.76 & 7.47 & +4.28 [+3.01, +5.49] \\
Qwen3-0.6B / rtn4 & 1500 & 1485 & 2.31 & 15.73 & 12.46 & +3.28 [+1.69, +4.81] \\
Llama-3.2-1B / nf4 & 1500 & 1485 & 0.87 & 7.29 & 3.77 & +3.52 [+2.58, +4.43] \\
Llama-3.2-1B / rtn4 & 1500 & 1485 & 1.55 & 12.14 & 7.27 & +4.87 [+3.60, +6.10] \\
Qwen3-1.7B / nf4 & 1500 & 1485 & 1.29 & 5.60 & 3.57 & +2.03 [+1.18, +2.85] \\
Qwen3-1.7B / rtn4 & 1500 & 1485 & 2.03 & 8.70 & 6.33 & +2.37 [+1.20, +3.45] \\
Llama-3.2-3B / nf4 & 1500 & 1485 & 1.23 & 5.17 & 2.09 & +3.09 [+2.35, +3.78] \\
Llama-3.2-3B / rtn4 & 1500 & 1485 & 1.35 & 5.66 & 2.69 & +2.97 [+2.17, +3.78] \\
Qwen3-4B / nf4 & 1500 & 1485 & 1.02 & 2.33 & 1.08 & +1.25 [+0.75, +1.76] \\
Qwen3-4B / rtn4 & 1500 & 1485 & 1.68 & 4.26 & 2.42 & +1.84 [+1.10, +2.54] \\
Qwen3-8B / nf4 & 1500 & 1485 & 1.00 & 2.13 & 1.21 & +0.92 [+0.42, +1.41] \\
Qwen3-8B / rtn4 & 1500 & 1485 & 1.31 & 2.95 & 1.35 & +1.60 [+1.02, +2.15] \\
GRM-gemma2-2B / nf4 & 1500 & 1485 & 0.48 & 6.04 & 3.43 & +2.60 [+1.75, +3.47] \\
GRM-gemma2-2B / rtn4 & 1500 & 1485 & 0.68 & 7.73 & 5.39 & +2.35 [+1.31, +3.33] \\
GRM-Llama3.2-3B / nf4 & 1500 & 1485 & 0.63 & 5.37 & 3.03 & +2.34 [+1.51, +3.16] \\
GRM-Llama3.2-3B / rtn4 & 1500 & 1485 & 0.84 & 6.89 & 4.51 & +2.37 [+1.32, +3.39] \\
\bottomrule
\end{tabular}

\end{table}

\begin{table}[H]
\centering\small
\setlength{\tabcolsep}{4pt}
\caption{\textbf{Decomposition of quantization damage.} Raw $\sig$ is in score units and differs
by architecture (Llama-3.2 heads emit smaller margins), so compare it within a family; occupancy
$P(z<2)$ is scale-free and is what predicts the flip rate. Nearly all flips fall inside the window,
and the model's prediction exceeds the observation in every row. The comparison set pools RewardBench with every correct-vs-trap comparison from the
adversarial pools for \nf{}, \rtn{} and \gpt{} ($n=6{,}765$), so those rates are diluted by the
wider-margin trap comparisons and sit below the RewardBench-only rates of Table~\ref{tab:churn}.
FP4 and all Qwen3-4B rows have no trap runs and use RewardBench only ($n=2{,}985$), which is why
their rates coincide with Table~\ref{tab:churn}.}
\label{tab:law}
\begin{tabular}{llccccc}
\toprule
& & & $P(z{<}2)$ & flips & obs. & pred. \\
Model & Quant. & $\sig$ & \% & $z{<}2$ \% & \% & \% \\
\midrule
Qwen3-0.6B & NF4 & 1.31 & 24.9 & 97.7 & 3.80 & 5.23 \\
 & FP4 & 2.27 & 64.1 & 95.6 & 12.26 & 15.06 \\
 & INT4-RTN & 2.18 & 37.1 & 94.2 & 7.36 & 8.23 \\
 & GPTQ & 1.32 & 25.0 & 98.2 & 4.17 & 5.27 \\
\midrule
Llama-3.2-1B & NF4 & 0.75 & 17.9 & 98.2 & 2.45 & 3.65 \\
 & FP4 & 1.28 & 45.6 & 98.6 & 6.97 & 10.29 \\
 & INT4-RTN & 1.28 & 27.1 & 97.4 & 4.02 & 5.95 \\
 & GPTQ & 1.21 & 26.0 & 88.9 & 5.17 & 5.66 \\
\midrule
Qwen3-1.7B & NF4 & 1.11 & 13.3 & 98.5 & 2.01 & 2.65 \\
 & FP4 & 2.00 & 39.6 & 96.9 & 5.43 & 8.07 \\
 & INT4-RTN & 1.77 & 21.4 & 95.3 & 3.81 & 4.26 \\
 & GPTQ & 1.43 & 17.5 & 97.5 & 2.37 & 3.44 \\
\midrule
Llama-3.2-3B & NF4 & 1.00 & 11.4 & 100.0 & 1.46 & 2.35 \\
 & FP4 & 1.38 & 27.0 & 96.2 & 3.55 & 5.44 \\
 & INT4-RTN & 1.24 & 13.9 & 96.9 & 1.88 & 2.86 \\
 & GPTQ & 0.91 & 10.4 & 94.2 & 1.52 & 2.14 \\
\midrule
Qwen3-4B & NF4 & 0.99 & 11.9 & 100.0 & 1.24 & 2.10 \\
 & FP4 & 1.57 & 20.8 & 98.2 & 1.91 & 3.69 \\
 & INT4-RTN & 1.65 & 21.9 & 98.5 & 2.28 & 3.91 \\
 & GPTQ & 0.96 & 11.6 & 92.5 & 1.78 & 2.03 \\
\midrule
Qwen3-8B & NF4 & 0.92 & 7.0 & 100.0 & 0.71 & 1.33 \\
 & FP4 & 1.23 & 14.6 & 97.6 & 1.41 & 2.57 \\
 & INT4-RTN & 1.10 & 8.6 & 97.2 & 1.05 & 1.62 \\
 & GPTQ & 0.65 & 4.7 & 88.9 & 0.67 & 0.91 \\
\bottomrule
\end{tabular}

\end{table}

\section{Cascade Frontier}
\label{app:frontier}

\begin{figure}[H]
\centering
\includegraphics[width=0.62\linewidth]{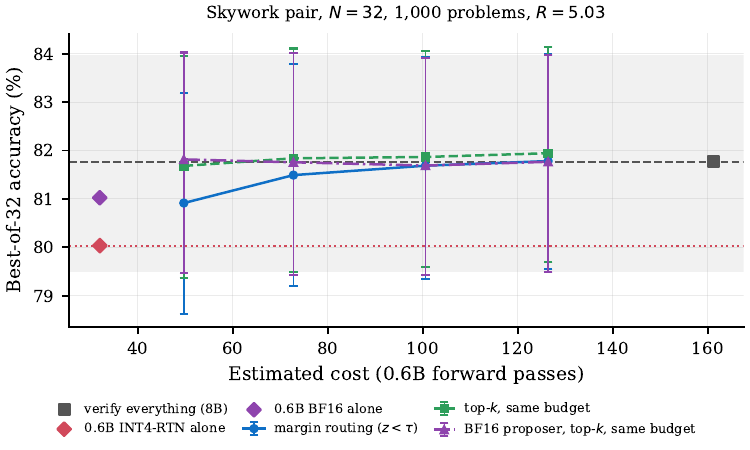}
\caption{\textbf{Accuracy--cost frontier for selective escalation}, Skywork pair, $N=32$,
1{,}000 problems, cost in proposer forward passes at the measured ratio $R=5.03$. The GRM pair is
omitted: its measured $R$ is $1.00$, so its cost axis is degenerate. Error bars are problem-level bootstrap intervals.}
\label{fig:cascade}
\end{figure}

Table~\ref{tab:cascade} tabulates the cascade and Fig.~\ref{fig:cascade} plots its accuracy--cost frontier.

\begin{table}[H]
\centering\small
\setlength{\tabcolsep}{3.5pt}
\caption{\textbf{Selective escalation}, $N=32$, 1{,}000 problems. ``Gap'': fraction of the gap
recovered; ``Save'': compute reduction at measured $R=5.03$. GRM's measured $R$ is $1.00$ (``--'').}
\label{tab:cascade}
\begin{tabular}{llccccc}
\toprule
& Prop. / Verif. & & Escal. & Acc. & Gap & Save \\
Cascade & acc.\ \% & $\tau$ & \% & \% & \% & \% \\
\midrule
0.6B-\rtn{} $\to$ 8B & 80.03 / 81.77 & 0.25 & 11.0 & 80.92 & 51 & 66.4 \\
 &  & 0.50 & 25.3 & 81.49 & 84 & 52.5 \\
 &  & 0.75 & 42.6 & 81.69 & 95 & 35.8 \\
 &  & 1.00 & 58.6 & 81.78 & 101 & 20.3 \\
\midrule
GRM-2B-\rtn{} $\to$ GRM-3B & 78.62 / 79.87 & 0.50 & 5.5 & 78.76 & 11 & -- \\
 &  & 1.00 & 9.4 & 79.05 & 35 & -- \\
 &  & 2.00 & 20.8 & 79.48 & 69 & -- \\
\bottomrule
\end{tabular}

\end{table}

\section{Audit Protocol and Selection-Rule Baselines}
\label{app:audit}
\label{app:fidelity}

The audit of \S\ref{sec:audit} is offline: it uses BF16 reference margins on held-out problems,
which an auditor has and a router does not. Within one (model, quantizer) dividing by a constant
$\sig$ cannot change any ranking, so the model's value can only appear in probability calibration
or in transfer across distributions, which is what Table~\ref{tab:audit} measures. Splits are by
problem throughout, and every seed and setting is reported.

\begin{table}[H]
\centering\small
\caption{\textbf{Held-out audit and cross-distribution transfer.} Mean absolute error in
predicting churn on data not used to fit the predictor; $n$ counts (setting $\times$ seed) rows,
and the last two columns count the rows in which the $\sig$ model's absolute error is the smaller
one. A paired bootstrap on the transfer rows (2{,}000 resamples of source \emph{and} target
problems, refitting $\sig$ and the logistic on each draw) puts the $\sig$-over-rate difference
entirely below zero in $16/19$ settings for RewardBench $\to$ traps (mean $-2.80$ pp) and $13/19$
for the reverse ($-2.25$ pp); against the logistic the same test gives $3/19$ and $2/19$ wins
against $8/19$ and $0/19$ losses. Per-setting values for every row are released in
\texttt{paper/audit\_per\_setting.csv}.}
\label{tab:audit}
\setlength{\tabcolsep}{4.5pt}
\begin{tabular}{lrcccrr}
\toprule
& & \multicolumn{3}{c}{mean $|$churn prediction error$|$ (pp)} & \multicolumn{2}{c}{$\sig$ wins} \\
\cmidrule(lr){3-5}\cmidrule(lr){6-7}
Setting & $n$ & $\sig$ model & rate & logistic & vs.\ rate & vs.\ logistic \\
\midrule
\multicolumn{7}{l}{\emph{In-distribution}: calibration and test problems from one dataset} \\
RewardBench, $n_{\mathrm{cal}}{=}200$ & 90 & 2.35 & 0.99 & 0.96 & 11/90 & 10/90 \\
RewardBench, $n_{\mathrm{cal}}{=}500$ & 90 & 2.29 & 0.73 & 0.67 & 5/90 & 6/90 \\
Traps, $n_{\mathrm{cal}}{=}200$ & 57 & 0.21 & 0.47 & 0.36 & 39/57 & 34/57 \\
Traps, $n_{\mathrm{cal}}{=}500$ & 57 & 0.23 & 0.27 & 0.22 & 34/57 & 30/57 \\
\midrule
\multicolumn{7}{l}{\emph{Transfer}: fit on one dataset, predict churn on the other (whole source)} \\
RewardBench $\to$ traps & 19 & 0.59 & 3.40 & 0.33 & 19/19 & 4/19 \\
Traps $\to$ RewardBench & 19 & 1.10 & 3.40 & 1.55 & 17/19 & 14/19 \\
\midrule
\multicolumn{7}{l}{\emph{Transfer, budget-matched}: source refit on $n_{\mathrm{cal}}$ problems} \\
RewardBench $\to$ traps, $n_{\mathrm{cal}}{=}200$ & 57 & 0.56 & 3.61 & 0.46 & 55/57 & 19/57 \\
RewardBench $\to$ traps, $n_{\mathrm{cal}}{=}500$ & 57 & 0.59 & 3.34 & 0.40 & 52/57 & 16/57 \\
Traps $\to$ RewardBench, $n_{\mathrm{cal}}{=}200$ & 57 & 1.06 & 3.60 & 2.28 & 51/57 & 44/57 \\
Traps $\to$ RewardBench, $n_{\mathrm{cal}}{=}500$ & 57 & 1.07 & 3.39 & 1.56 & 49/57 & 39/57 \\
\bottomrule
\end{tabular}

\end{table}

\paragraph{Row counts, calibration budgets, and metric definitions.}
Thirty (model, quantizer) settings have RewardBench scores and $19$ have both RewardBench and
traps, so the in-distribution blocks contain $30\times3$ and $19\times3$ seed-split rows. The
whole-source transfer block draws no split (it fits on every source comparison), so it
contains one row per setting, $19$. In that block $\sig$ is the standard deviation of
$m_{\text{quant}}-m_{\text{BF16}}$ over all source comparisons, the rate baseline is the source
churn rate, and the logistic is fit by maximum likelihood on all source comparisons; all three are
then applied unchanged to every target comparison. Sources are $2{,}985$ RewardBench comparisons
(one per problem) and $3{,}780$ trap comparisons ($1{,}260$ problems $\times$ 3). Because that is a
larger budget than the $200$/$500$ problems the in-distribution rows use, the budget-matched block
repeats the whole procedure on a random $n_{\text{cal}}$-problem subsample of the source, three
seeds, and reproduces the full-source numbers to within $0.04$ pp. ECE uses $10$ equal-width bins
on $[0,1]$, half-open, weighted by bin occupancy, with empty bins contributing zero. For the
constant-rate baseline every prediction falls in one bin, so its ECE equals its absolute aggregate
error exactly (verified to machine precision); ECE is therefore not evidence independent of the
error column for that baseline, which is why the discrimination comparison in \S\ref{sec:audit} is
made with the Brier score instead.

\paragraph{What $\Phi(-\tau)$ does and does not say.}
Eq.~\eqref{eq:law} is stated on the latent $z$; routing conditions on the observed $\hat z$ ($c^\star$
is an empirical maximum, so $\hat z$ is biased upward), and we make no analytic claim on $\hat z$.
Empirically, unescalated disagreement stays under $\Phi(-\tau)$ only \emph{(i)} against $M_q$'s
\emph{own BF16}; \emph{(ii)} for scalar reward models, not judges; \emph{(iii)} for the zero-mean
model.

\paragraph{Cost model (an estimate).}
Cost is in proposer passes; $R=5.03$ is the measured scoring-throughput ratio of 8B to
0.6B-\rtn{} at our batch sizes (the parameter ratio $13.3$ is a FLOP bound batching does not
realize). Estimated saving is $1-1/R-(1+f(N-1))/N$; we did not time the full cascade. The GRM
pair's $R$ is $1.00$, so it cannot save compute.

Table~\ref{tab:fidelity} reports the baselines described in \S\ref{sec:cascade}, and
Table~\ref{tab:controls} the two proposer controls. On the looseness of the envelope, measured against the proposer's \emph{own} BF16 (the only
comparison it is scoped to): at $\tau=0.5$ the nominal $\Phi(-0.5)=0.31$ compares to a realized
disagreement of $0.23$ among unescalated comparisons.

\begin{table}[H]
\centering\small
\caption{\textbf{Fidelity to the full-verifier decision.} The fraction of problems on which the
cascade selects exactly what verifying every candidate would, at matched escalation budget. Margin
routing is far better than random, but slightly worse than escalating the proposer's top-$k$. What
the margin rule adds over top-$k$ is an empirical envelope on disagreement, not selection
quality.}
\label{tab:fidelity}
\begin{tabular}{cccccc}
\toprule
$N$ & $\tau$ & budget \% & margin \% & top-$k$ \% & random \% \\
\midrule
32 & 0.25 & 11.0 & 33.8 & 37.8 & 24.6 \\
32 & 0.50 & 25.3 & 56.1 & 59.5 & 36.5 \\
32 & 1.00 & 58.6 & 86.1 & 87.0 & 65.1 \\
64 & 0.25 & 6.8 & 26.8 & 30.5 & 16.7 \\
64 & 0.50 & 18.7 & 50.3 & 54.5 & 27.4 \\
64 & 1.00 & 51.7 & 82.8 & 84.2 & 57.1 \\
\bottomrule
\end{tabular}

\end{table}

\begin{table}[H]
\centering\footnotesize
\setlength{\tabcolsep}{3pt}
\caption{\textbf{Proposer controls}, $N=32$, 1{,}000 problems, routed with the \rtn{}
proposer's $\sig$. Budgets differ by proposer at a given $\tau$, so the matched comparison is in
Table~\ref{tab:paired}; there the escalated \emph{fraction} is matched, and because the BF16 and
\rtn{} 0.6B proposers score at the same measured rate ($65$ comparisons/s) this also matches total
estimated cost. Rows with different proposers at a fixed $\tau$ match neither.
Escalating the quantized proposer to its \emph{own} BF16 isolates quantization damage.}
\label{tab:controls}
\resizebox{\linewidth}{!}{\begin{tabular}{llccccc}
\toprule
Cascade & Prop. / Verif. \% & $\tau$ & Escal. \% & Acc. \% & Gap \% & Fidelity \% \\
\midrule
0.6B-\rtn{} $\to$ 0.6B-BF16 (same model) & 80.03 / 81.02 & 0.25 & 11.0 & 80.67 & 64 & 47.0 \\
 &  & 0.50 & 25.3 & 80.82 & 80 & 70.4 \\
 &  & 1.00 & 58.6 & 80.88 & 86 & 94.2 \\
\midrule
0.6B-BF16 $\to$ 8B-BF16 (no quantization) & 81.02 / 81.77 & 0.25 & 15.3 & 81.71 & 92 & 57.0 \\
 &  & 0.50 & 34.8 & 81.76 & 99 & 80.5 \\
 &  & 1.00 & 67.3 & 81.56 & 72 & 96.3 \\
\bottomrule
\end{tabular}
}
\end{table}

\begin{table}[H]
\centering\small
\caption{\textbf{Problem-level paired differences}, $N=32$, 1{,}000 problems, 2{,}000 bootstrap
resamples of problems. All arms are evaluated on the same 40 candidate draws per problem; the
``same budget'' rows match escalated fraction, hence estimated cost.}
\label{tab:paired}
\begin{tabular}{lccc}
\toprule
Comparison ($N=32$) & $\Delta$ acc.\ (pt) & 95\% CI & escalated \% \\
\midrule
8B verifier $-$ 0.6B \rtn{} (total gap) & +1.73 & [+0.69, +2.88] & -- \\
0.6B BF16 $-$ 0.6B \rtn{} (quantization) & +0.99 & [+0.07, +1.96] & -- \\
8B verifier $-$ 0.6B BF16 (capacity) & +0.74 & [-0.21, +1.73] & -- \\
\midrule
margin $\tau{=}0.5$ $-$ proposer & +1.46 & [+0.71, +2.30] & 25.3 \\
margin $-$ top-$k$, same budget & -0.35 & [-0.71, -0.00] & 25.3 \\
margin $-$ BF16-proposer top-$k$, same budget & -0.26 & [-0.80, +0.30] & 25.3 \\
\rtn{}$\to$own BF16 $\tau{=}0.5$ $-$ proposer & +0.79 & [+0.12, +1.52] & 25.3 \\
\bottomrule
\end{tabular}

\end{table}

\section{Adversarial and Amplification Figures}
\label{app:harmfig}

\begin{figure}[H]
\centering
\includegraphics[width=\linewidth]{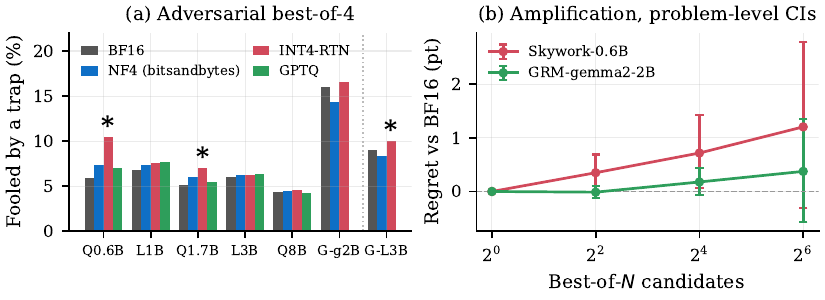}
\caption{\textbf{(a)} Adversarial fooling rate by model and quantizer, both families (dotted line
separates them); $*$ marks $p<0.05$ for \rtn{}. \textbf{(b)} Best-of-$N$ regret against the BF16
selector on the 1{,}000-problem pool, with problem-level bootstrap intervals. Point estimates
increase with $N$ in both families, but only $N=4$ and $N=16$ for Skywork-0.6B exclude zero.}
\label{fig:harm}
\end{figure}

Fig.~\ref{fig:harm} visualizes the trap and amplification results of \S\ref{sec:harm}.

\section{Validating the GPTQ Implementation}
\label{app:gptqval}

\gpt{} minimizes \emph{activation-weighted} reconstruction error $\|(W-\hat W)X\|$ and deliberately
accepts larger raw weight error to do so, so weight MSE is not a valid check. We capture real
activations $X$ on 32 RewardBench prompts (used only to measure error, not for calibration) and
compare against same-bit-width \rtn{} on the first eight quantized linear layers:

\begin{center}\small
\begin{tabular}{lcc}
\toprule
Model & layers where \gpt{} $<$ \rtn{} & median error ratio \\
\midrule
Skywork-Reward-V2-Qwen3-0.6B    & 8/8 & 0.295 \\
Skywork-Reward-V2-Llama-3.2-1B  & 8/8 & 0.329 \\
Skywork-Reward-V2-Llama-3.2-3B  & 8/8 & 0.204 \\
\bottomrule
\end{tabular}
\end{center}

\gpt{} attains three to five times lower activation-weighted error than \rtn{} in every layer
tested, so the implementation performs the optimization it is supposed to. This matters for
interpreting Llama-3.2-1B, where \gpt{} nonetheless produces \emph{more} decision churn than
\rtn{}: better reconstruction, worse decisions.

\section{Downstream Protocol and Per-Dataset Results}
\label{app:downstream}

The pool is 1{,}000 problems with 64 candidates each; an earlier 200-problem pool at $N{=}256$
proved too small to resolve sub-2-point downstream effects, and we report only the larger one.
For $N<64$ we draw $N$ distinct candidates per problem, 40 times, giving both selectors the
\emph{same} draws; at $N=64$ there is one possible subset, so a single deterministic draw is used
and repeated draws would add no observations. Ties in the arg-max are broken by lowest candidate
index, identically for both selectors; BF16 score precision leaves $32$--$39$ distinct values per
problem, so the top score is tied on $12$--$15\%$ of problems. Per-problem outcomes are averaged
over draws, then whole problems are resampled ($3{,}000$ times) stratified by dataset, preserving
the MATH/GSM8K mixture; ratios such as the judge preference are recomputed inside each resample.
The sign convention is $\Delta = \text{BF16} - \text{NF4}$ throughout. Candidate texts are
$99.9\%$ unique, so identity disagreement is not an artifact of duplicate candidates. Per-problem
records are released alongside the summary table.

The six-pair replication of Table~\ref{tab:matrix} runs this same code: \texttt{downstream\_matrix.py}
imports the protocol and varies only which score files are loaded, so the headline pair's numbers
are reproduced to the last digit. Every quantized model is compared against \emph{its own} BF16
reference on the same pools; the arbiter is Skywork-8B BF16 throughout, which is same-family for the
Skywork pairs and cross-family for the GRM pairs. A \emph{tied-max} switch is one where at least one
selector's maximum was not unique, so the two selectors were separated by the tie rule rather than by
a preference; those are $20$--$31\%$ of all switches, which is why we report the untied rate as well.

\begin{table}[H]
\centering\footnotesize
\setlength{\tabcolsep}{4pt}
\caption{\textbf{The downstream result replicates across families and quantizers.} Six (model,
quantizer) pairs, each against its own BF16 reference, identical pools, draws and tie rule.
Selections change far more often than answers in every pair, by $5.3$--$7.7\times$ at $N=64$,
and only $4$ of the $24$ accuracy intervals (6 pairs $\times$ 4 values of $N$) exclude zero.
$\Delta$acc.\ is BF16 minus quantized, so positive favours BF16.}
\label{tab:matrix}
\resizebox{\linewidth}{!}{\begin{tabular}{llrrrrr}
\toprule
& & \multicolumn{3}{c}{different selection (\% of problems)} & \multicolumn{2}{c}{consequence} \\
\cmidrule(lr){3-5}\cmidrule(lr){6-7}
Model / quantizer & $N$ & all & untied & at a tied max & diff.\ answer \% & $\Delta$acc.\ (pp) \\
\midrule
Skywork-V2-Llama-3.2-1B / \nf{} & 16 & 39.8 & 28.5 & 28.4 & 4.1 & +0.35 [-0.01, +0.71] \\
 & 64 & 50.3 & 34.5 & 31.4 & 4.5 & +0.80 [-0.10, +1.70] \\
\addlinespace
Skywork-V2-Llama-3.2-1B / \gpt{} & 16 & 49.7 & 37.9 & 23.7 & 5.8 & +0.28 [-0.19, +0.76] \\
 & 64 & 61.4 & 42.7 & 30.5 & 6.6 & +0.70 [-0.50, +1.90] \\
\addlinespace
Skywork-V2-Qwen3-0.6B / \nf{} & 16 & 50.8 & 39.6 & 22.1 & 4.6 & +0.44 [+0.01, +0.88] \\
 & 64 & 66.2 & 47.7 & 27.9 & 7.1 & +0.90 [-0.30, +2.20] \\
\addlinespace
Skywork-V2-Qwen3-0.6B / \rtn{} & 16 & 70.0 & 59.9 & 14.4 & 7.6 & +0.62 [-0.07, +1.33] \\
 & 64 & 82.1 & 65.7 & 20.0 & 9.6 & +1.20 [-0.40, +2.80] \\
\addlinespace
GRM-gemma2-2B / \nf{} & 16 & 16.5 & 12.1 & 26.9 & 1.9 & +0.06 [-0.13, +0.26] \\
 & 64 & 21.1 & 14.6 & 30.8 & 2.6 & -0.10 [-0.80, +0.70] \\
\addlinespace
GRM-gemma2-2B / \rtn{} & 16 & 21.8 & 17.4 & 20.0 & 2.7 & +0.22 [-0.03, +0.47] \\
 & 64 & 28.9 & 22.1 & 23.5 & 4.2 & +0.40 [-0.50, +1.40] \\
\bottomrule
\end{tabular}
}
\end{table}

\begin{table}[H]
\centering\small
\caption{\textbf{What changes downstream.} Llama-3.2-1B at BF16 vs.\ \nf{} ($-0.03$ RewardBench
points, $p=1.00$) selecting best-of-$N$ over identical candidates, 1{,}000 problems. $^\dagger$
conditional on the two selecting different candidates. ``judge'' is Skywork-8B BF16, a larger
evaluator of the same family, not an independent assessor, on disagreements only. Selections
diverge sharply; the returned \emph{answer} usually does not.}
\label{tab:ship}
\begin{tabular}{ccccccc}
\toprule
& switches & \multicolumn{3}{c}{of switches (\%), sums to 100} & & judge \\
\cmidrule(lr){3-5}
$N$ & \% of prob. & same answer & diff.\ answer & no parse & $\Delta$acc.\ (pp) [95\% CI] & \% \\
\midrule
8 & 32.6 & 87.9 & 10.6 & 1.5 & +0.24 [-0.03, +0.53] & 51.3 \\
16 & 39.8 & 88.9 & 10.2 & 0.9 & +0.35 [-0.01, +0.71] & 50.4 \\
32 & 45.8 & 89.6 & 9.8 & 0.6 & +0.22 [-0.26, +0.73] & 49.9 \\
64 & 50.3 & 90.7 & 8.9 & 0.4 & +0.80 [-0.10, +1.70] & 49.7 \\
\bottomrule
\end{tabular}

\end{table}

\begin{table}[H]
\centering\small
\caption{\textbf{Downstream results by dataset.} Answer-level divergence is roughly three times
higher on MATH-500 than on GSM8K, so the effect is distribution-dependent.}
\label{tab:shipds}
\begin{tabular}{llccccc}
\toprule
$N$ & dataset & id \% & answer \% & same answer$^\dagger$ \% & $\Delta$acc.\ (pp) & judge \% \\
\midrule
8 & MATH-500 & 31.6 & 4.8 & 81.8 & +0.60 & 51.8 \\
8 & GSM8K & 33.6 & 2.2 & 93.6 & -0.11 & 50.8 \\
32 & MATH-500 & 45.4 & 6.8 & 83.9 & +0.44 & 50.1 \\
32 & GSM8K & 46.2 & 2.2 & 95.2 & -0.00 & 49.6 \\
64 & MATH-500 & 50.2 & 7.2 & 84.9 & +0.60 & 49.4 \\
64 & GSM8K & 50.4 & 1.8 & 96.4 & +1.00 & 50.0 \\
\bottomrule
\end{tabular}

\end{table}

\section{Generative Judges}
\label{app:judge}

Tables~\ref{tab:judge} and \ref{tab:jtrap} give the judge results summarized in \S\ref{sec:law}
and \S\ref{sec:harm}. A judge's margin is meaningful only if it actually emits a verdict token: on
200 held-out prompts the arg-max token is ``A'' or ``B'' for $100\%$, $98\%$ and $100\%$ of the
three judges. Qwen2.5-7B under \nf{} yields harm:help of exactly 73:35, the same integers as the
pooled Skywork \nf{} row; the two are computed independently and the coincidence is numerical.

\begin{table}[H]
\centering\small
\setlength{\tabcolsep}{4pt}
\caption{\textbf{Generative pairwise judges on RewardBench} ($n=2{,}985$). The margin is the
difference of the ``A'' and ``B'' verdict logits, averaged over both response orderings to remove
position bias. Churn tracks window occupancy exactly as it does for scalar reward models, but
$\dagger$ marks the two settings where $\Phi(-z)$ \emph{under}-predicts the observed flip rate;
the bound is not conservative here, which is why the envelope is scoped to scalar models.}
\label{tab:judge}
\begin{tabular}{llcccccc}
\toprule
Judge & Quant. & BF16 acc.\ \% & $\Delta$acc. & Flip \% & $\sig$ & $P(z{<}2)$ \% & pred.\ \% \\
\midrule
Qwen2.5-7B-Instruct & NF4 & 83.99 & -0.84 & 4.79 & 2.15 & 22.1 & 4.34$^{\dagger}$ \\
 & INT4-RTN & 83.99 & +0.07 & 6.37 & 3.36 & 33.3 & 6.77 \\
\midrule
Mistral-7B-Instruct-v0.2 & NF4 & 72.56 & -0.94 & 6.97 & 1.58 & 48.0 & 11.24 \\
 & INT4-RTN & 72.56 & -2.35 & 7.97 & 1.96 & 54.5 & 13.25 \\
\midrule
Qwen2.5-3B-Instruct & NF4 & 67.40 & +2.71 & 16.98 & 4.38 & 80.7 & 20.11 \\
 & INT4-RTN & 67.40 & -11.42 & 37.55 & 5.58 & 89.9 & 23.58$^{\dagger}$ \\
\bottomrule
\end{tabular}

\end{table}

\begin{table}[H]
\centering\small
\setlength{\tabcolsep}{2.8pt}
\caption{\textbf{Generative judges on the adversarial traps} ($n=3{,}780$ correct-vs-trap
comparisons per judge, both pools). ``Prefers trap'' is the fraction of comparisons on which the
judge ranked the verified-wrong candidate above the verified-correct one. The two 7B judges move in opposite directions under the same quantizer. Qwen2.5-3B was excluded:
its BF16 trap error is too high for a fooling-rate comparison to be meaningful. The $p$ column is an
unclustered McNemar test shown for continuity; problem-level clustered bootstrap CIs on harmful$-$helpful: Qwen2.5/NF4 $[18, 58]$, Qwen2.5/RTN4 $[1, 40]$, Mistral/NF4 $[-90, -39]$, Mistral/RTN4 $[-101, -47]$, which are the values we rely on.}
\label{tab:jtrap}
\begin{tabular}{llccccc}
\toprule
Judge & Quant. & BF16 prefers trap \% & Quant. \% & harm:help & $p$ & shift$/\sigma$ \\
\midrule
Qwen2.5-7B-Instruct & NF4 & 4.87 & 5.87 & 73:35 & \textbf{$<$0.001} & -0.90 \\
 & INT4-RTN & 4.87 & 5.40 & 64:44 & 0.07 & -0.80 \\
\midrule
Mistral-7B-Instruct-v0.2 & NF4 & 10.71 & 9.02 & 61:125 & \textbf{$<$0.001} & +0.37 \\
 & INT4-RTN & 10.71 & 8.76 & 66:140 & \textbf{$<$0.001} & +0.16 \\
\bottomrule
\end{tabular}

\end{table}

\section{Systematic Bias by Quantizer}
\label{app:bias}

Table~\ref{tab:bias} gives the mean shifts discussed in \S\ref{sec:harm};
Table~\ref{tab:contract} the regression separating contraction (slope) from an additive shift
(intercept); Table~\ref{tab:threeparam} the held-out test of the three-parameter model
$\tilde m=c\,m+b+\eta$ against the one-parameter model, on harmful and helpful flip counts and
log-likelihood, with the residual's excess kurtosis; and Table~\ref{tab:localbase} the in-window
base rates and per-comparison flip rates by original correctness. The pooled harm:help results in
the main text use the problem-level clustered bootstrap described in \S\ref{sec:setup}: the
pooled Skywork \nf{}, \rtn{} and \gpt{} results and the GRM \nf{} reversal all have intervals
excluding zero ($p\le0.002$); GRM \rtn{} is marginal ($p=0.031$, CI $[2, 38]$).

\begin{table}[H]
\centering\small
\setlength{\tabcolsep}{3.5pt}
\caption{\textbf{Three-parameter model, held-out.} $c,b,\sigma_\eta$ fit on even-indexed
comparisons; harmful (originally-correct) and helpful (originally-incorrect) flips on odd-indexed
comparisons, observed / one-parameter prediction / three-parameter prediction; $\Delta\ell\ell$ is
the held-out log-likelihood gain of the three-parameter model (positive favours it). It wins in
$17/30$ settings and is worse for \rtn{} in five of eight; splits are grouped by problem.}
\label{tab:threeparam}
\begin{tabular}{lcccccc}
\toprule
Model / quant. & $c$ & $b/\sigma_\eta$ & harm obs / 1p / 3p & help obs / 1p / 3p & $\Delta\ell\ell$ & kurt. \\
\midrule
Qwen3-0.6B / nf4 & 0.87 & +0.22 & 82 / 133 / 104 & 58 / 57 / 74 & +10.0 & 2.4 \\
Qwen3-0.6B / rtn4 & 0.62 & -0.09 & 171 / 216 / 268 & 77 / 75 / 72 & -12.4 & 0.6 \\
Llama-3.2-1B / nf4 & 0.95 & +0.14 & 41 / 86 / 71 & 49 / 42 / 49 & +10.3 & 1.5 \\
Llama-3.2-1B / rtn4 & 0.83 & +0.12 & 86 / 150 / 124 & 56 / 57 / 64 & +10.9 & 1.2 \\
Qwen3-1.7B / nf4 & 0.92 & -0.04 & 50 / 69 / 73 & 27 / 32 / 30 & -1.5 & 2.0 \\
Qwen3-1.7B / rtn4 & 0.82 & -0.23 & 105 / 117 / 162 & 34 / 43 / 34 & -12.1 & 1.0 \\
Llama-3.2-3B / nf4 & 0.95 & +0.37 & 34 / 61 / 35 & 15 / 25 / 35 & +1.4 & 4.6 \\
Llama-3.2-3B / rtn4 & 0.94 & -0.11 & 38 / 76 / 86 & 28 / 28 / 25 & -7.8 & 2.4 \\
Qwen3-4B / nf4 & 1.03 & -0.41 & 12 / 26 / 41 & 4 / 10 / 5 & -10.7 & 2.5 \\
Qwen3-4B / rtn4 & 0.88 & +0.05 & 22 / 51 / 42 & 14 / 14 / 14 & +6.4 & 4.9 \\
Qwen3-8B / nf4 & 1.00 & +0.55 & 14 / 33 / 14 & 15 / 18 / 31 & +0.4 & 2.2 \\
Qwen3-8B / rtn4 & 0.95 & +0.07 & 25 / 41 / 37 & 15 / 21 / 22 & +1.2 & 2.0 \\
GRM-gemma2-2B / nf4 & 0.98 & +0.10 & 35 / 68 / 61 & 40 / 45 / 50 & +3.2 & 3.8 \\
GRM-gemma2-2B / rtn4 & 0.94 & -0.28 & 68 / 101 / 141 & 54 / 57 / 41 & -28.0 & 1.5 \\
GRM-Llama3.2-3B / nf4 & 0.95 & +0.01 & 35 / 76 / 72 & 29 / 35 / 35 & +3.7 & 0.6 \\
GRM-Llama3.2-3B / rtn4 & 0.95 & -0.10 & 64 / 103 / 117 & 42 / 42 / 38 & -8.1 & 0.6 \\
\bottomrule
\end{tabular}

\end{table}

\begin{table}[H]
\centering\footnotesize
\setlength{\tabcolsep}{2.6pt}
\caption{\textbf{In-window base rates.} Among comparisons with $z<2$: the share originally
correct, the harmful:helpful counts, the per-comparison flip rate of originally-correct and
originally-incorrect decisions, and the one-parameter model's over-prediction factor on the
harmful side. Incorrect decisions flip more often per comparison; the count asymmetry is base
rates.}
\label{tab:localbase}
\resizebox{\linewidth}{!}{\begin{tabular}{lcccccc}
\toprule
Model / quant. & $n_{z<2}$ & correct \% & harm:help & correct flip \% & incorrect flip \% & harm over-pred. \\
\midrule
Qwen3-0.6B / nf4 & 1682 & 77.1 & 164:87 & 12.7 & 22.5 & $\times$1.62 \\
Qwen3-0.6B / rtn4 & 2511 & 82.2 & 332:137 & 16.1 & 30.7 & $\times$1.26 \\
Llama-3.2-1B / nf4 & 1208 & 75.0 & 86:77 & 9.5 & 25.5 & $\times$2.09 \\
Llama-3.2-1B / rtn4 & 1831 & 79.7 & 159:106 & 10.9 & 28.6 & $\times$1.75 \\
Qwen3-1.7B / nf4 & 900 & 76.1 & 91:43 & 13.3 & 20.0 & $\times$1.38 \\
Qwen3-1.7B / rtn4 & 1449 & 81.6 & 183:63 & 15.5 & 23.6 & $\times$1.11 \\
Llama-3.2-3B / nf4 & 768 & 76.8 & 65:34 & 11.0 & 19.1 & $\times$1.79 \\
Llama-3.2-3B / rtn4 & 941 & 78.3 & 68:55 & 9.2 & 27.0 & $\times$2.00 \\
Qwen3-4B / nf4 & 355 & 78.6 & 25:12 & 9.0 & 15.8 & $\times$2.14 \\
Qwen3-4B / rtn4 & 654 & 84.4 & 42:25 & 7.6 & 24.5 & $\times$2.34 \\
Qwen3-8B / nf4 & 471 & 73.0 & 27:21 & 7.8 & 16.5 & $\times$2.34 \\
Qwen3-8B / rtn4 & 584 & 76.2 & 49:20 & 11.0 & 14.4 & $\times$1.63 \\
GRM-gemma2-2B / nf4 & 968 & 68.3 & 59:65 & 8.9 & 21.2 & $\times$1.96 \\
GRM-gemma2-2B / rtn4 & 1353 & 70.8 & 126:94 & 13.2 & 23.8 & $\times$1.49 \\
GRM-Llama3.2-3B / nf4 & 988 & 76.2 & 57:55 & 7.6 & 23.4 & $\times$2.18 \\
GRM-Llama3.2-3B / rtn4 & 1232 & 77.8 & 93:66 & 9.7 & 24.1 & $\times$1.61 \\
\bottomrule
\end{tabular}
}
\end{table}

\begin{table}[H]
\centering\small
\caption{\textbf{Contraction versus additive shift.} Linear regression of $\Delta m = m_{\text{quant}}
- m_{\text{BF16}}$ on $m_{\text{BF16}}$ over the trap comparisons. A negative slope is
scale contraction, which lowers positive margins without reversing any; the intercept is the
additive component. \rtn{}'s large negative mean shift is almost entirely slope; \nf{} carries a
small positive intercept.}
\label{tab:contract}
\begin{tabular}{lcccc}
\toprule
Model / quantizer & slope & intercept$/\sigma$ & mean shift$/\sigma$ & $R^2$ \\
\midrule
Qwen3-0.6B / nf4 & -0.09 & +0.37 & -0.28 & 0.08 \\
Qwen3-0.6B / rtn4 & -0.41 & +0.02 & -1.47 & 0.42 \\
Llama-3.2-1B / nf4 & -0.07 & +0.28 & -0.31 & 0.07 \\
Llama-3.2-1B / rtn4 & -0.17 & +0.20 & -0.76 & 0.20 \\
Qwen3-1.7B / nf4 & -0.09 & +0.05 & -0.78 & 0.12 \\
Qwen3-1.7B / rtn4 & -0.20 & -0.12 & -1.27 & 0.23 \\
Llama-3.2-3B / nf4 & -0.01 & +0.25 & +0.09 & 0.00 \\
Llama-3.2-3B / rtn4 & -0.09 & -0.12 & -1.04 & 0.15 \\
Qwen3-8B / nf4 & +0.05 & +0.49 & +1.38 & 0.13 \\
Qwen3-8B / rtn4 & -0.03 & +0.05 & -0.29 & 0.02 \\
GRM-gemma2-2B / nf4 & -0.02 & +0.41 & +0.12 & 0.04 \\
GRM-gemma2-2B / rtn4 & -0.10 & -0.21 & -0.91 & 0.21 \\
GRM-Llama3.2-3B / nf4 & -0.07 & +0.26 & -0.63 & 0.20 \\
GRM-Llama3.2-3B / rtn4 & -0.05 & -0.13 & -0.67 & 0.07 \\
\bottomrule
\end{tabular}

\end{table}

\begin{table}[H]
\centering\small
\caption{\textbf{Mean shift of the correct-vs-trap margin}, in units of $\sig$. A negative
value lowers positive margins but, as Table~\ref{tab:contract} shows, is mostly scale contraction
rather than an additive tilt; it should not be read as bias against the correct answer. \rtn{}
is negative in all seven models; \nf{}'s shift is smaller and changes sign by model.}
\label{tab:bias}
\begin{tabular}{lccc}
\toprule
Model & \nf{} & \rtn{} & \gpt{} \\
\midrule
Qwen3-0.6B & -0.28 & -1.47 & -1.00 \\
Llama-3.2-1B & -0.31 & -0.76 & -0.16 \\
Qwen3-1.7B & -0.78 & -1.27 & -0.23 \\
Llama-3.2-3B & +0.09 & -1.04 & -0.07 \\
Qwen3-8B & +1.38 & -0.29 & +0.08 \\
GRM-gemma2-2B & +0.12 & -0.91 & -- \\
GRM-Llama3.2-3B & -0.63 & -0.67 & -- \\
\bottomrule
\end{tabular}

\end{table}

\section{Independent-Family Replication}
\label{app:repl}

Table~\ref{tab:repl} reports the RewardBench decomposition for the two GRM models; the
corresponding trap results appear in Table~\ref{tab:adv} and the systematic shifts in
Table~\ref{tab:bias}.

\begin{table}[H]
\centering\small
\caption{\textbf{The model replicates on an independent reward model family.} Two GRM models
\citep{yang2024grm}, trained with a different recipe from Skywork-Reward-V2 (GRM-Llama-3.2-3B
shares only the base backbone), on
RewardBench ($n=2{,}985$; these runs have no adversarial trap pools, so the comparison set is
RewardBench only). ``pred.'' is $\mathbb{E}[\Phi(-z)]$, which exceeds the observed flip rate in
every row, as it does for every Skywork setting.}
\label{tab:repl}
\setlength{\tabcolsep}{3.5pt}
\begin{tabular}{llcccccc}
\toprule
Model & Quantizer & $\Delta$acc. (pt) & Flip \% & $\sig$ & $P(z{<}2)$ \% & flips $z{<}2$ \% & pred.\ \% \\
\midrule
GRM-gemma2-2B & NF4 & -0.60 & 2.95 & 0.49 & 25.7 & 97.7 & 5.94 \\
 & FP4 & -1.47 & 4.56 & 0.69 & 32.0 & 100.0 & 7.72 \\
 & INT4-RTN & -0.84 & 4.99 & 0.66 & 31.5 & 97.3 & 7.48 \\
\midrule
GRM-Llama3.2-3B & NF4 & -0.34 & 2.68 & 0.64 & 24.0 & 96.2 & 4.98 \\
 & FP4 & -0.40 & 3.62 & 0.83 & 29.3 & 96.3 & 6.30 \\
 & INT4-RTN & -0.57 & 3.85 & 0.85 & 29.7 & 95.7 & 6.46 \\
\bottomrule
\end{tabular}

\end{table}

\section{Claim-to-Evidence Table}
\label{app:checklist}

\begin{table}[H]
\centering\footnotesize
\caption{Every headline claim, the population it is measured on, where the evidence appears, and
how uncertainty is quantified.}
\label{tab:checklist}
\resizebox{\linewidth}{!}{\begin{tabular}{p{0.27\linewidth}p{0.22\linewidth}p{0.15\linewidth}p{0.28\linewidth}}
\toprule
Claim & Population & Evidence & Uncertainty \\
\midrule
Accuracy can move little while churn is large & RewardBench, 2{,}985 pairs; 30 model$\times$quantizer settings (not a full $8\times4$ grid) & Table~\ref{tab:churn} & bootstrap 95\% CI; exact McNemar, BH-corrected \\
Flips concentrate at low margin & pooled RB + traps, 6{,}765 comp./setting & Table~\ref{tab:law}, Fig.~\ref{fig:law}a & Wilson 95\% CI per bin \\
$\mathbb{E}[\Phi(-z)]$ exceeds the observed \emph{aggregate} rate & held-out odd-indexed RB pairs, 30 settings & Table~\ref{tab:heldout} & bootstrap CI on the difference; not a conditional bound \\
Relative noise and exposure fall with size & RB, within architecture family & Fig.~\ref{fig:mechanism} & point estimates; no CI \\
Benign best-of-16 harm is within noise & 2{,}319 benign problems, 4 RMs & \S\ref{sec:harm} & exact McNemar per model \\
Amplification: point-estimate trend only & 1{,}000 problems, 2 RMs & Fig.~\ref{fig:harm}b & problem bootstrap; CI excludes 0 only at $N{=}4,16$ for 0.6B \\
\rtn{} raises trap fooling on both families & 1{,}260 traps, 6 RMs & Table~\ref{tab:adv} & problem-level clustered bootstrap \\
\nf{} reverses direction between families & 1{,}260 traps & Table~\ref{tab:adv} & clustered bootstrap CIs exclude 0 both ways \\
Harm asymmetry is base rates & in-window comparisons, 8 RMs & Table~\ref{tab:localbase} & descriptive counts and rates \\
Adding contraction does not improve held-out prediction & held-out grouped split, 30 settings & Table~\ref{tab:threeparam} & held-out log-likelihood; wins $17/30$ \\
Selector returns a different output on $50.3\%$ of problems at $N{=}64$ & 1{,}000 problems, best-of-$N$ & Table~\ref{tab:ship} & stratified problem bootstrap \\
$90.7\%$ of those switches return the same final answer & same & Table~\ref{tab:ship} & answer-level disagreement $4.5\%$ \\
Downstream accuracy difference unresolved & same & Table~\ref{tab:ship} & $+0.80$ pp, CI $[-0.10, 1.70]$ \\
Escalation recovers most quantization damage & 1{,}000 problems, $N=32$ & Table~\ref{tab:paired} & problem-level paired bootstrap \\
Top-$k$ matches or beats margin routing & 1{,}000 problems, matched budget & Tables~\ref{tab:fidelity},~\ref{tab:paired} & paired bootstrap; CI excludes 0 \\
Audit: direct churn estimation wins on RewardBench & 3 seeds $\times$ 2 calib.\ sizes, grouped & Table~\ref{tab:audit} & per-setting errors; $\sig$ wins $11/90$ \\
Margin models lead in cross-dataset transfer & 2 directions, 2 datasets & Table~\ref{tab:audit} & not 19 independent domain shifts \\
Churn extends to generative judges & RewardBench, 3 judges & Table~\ref{tab:judge} & bootstrap CI; envelope fails in 2/6 \\
\bottomrule
\end{tabular}
}
\end{table}

\section{Negative Results in Full}
\label{app:negative}

\begin{table}[H]
\centering\small
\caption{\textbf{Adversarial best-of-4 fooling rates} ($n=1{,}260$ problems; each has one verified
correct and three verified-wrong but persuasive candidates). ``Fooled'' means the model ranked a
verifiably wrong solution first. Pooled rows aggregate each family; $\dagger$ marks tests that are
significant at raw $p<0.05$ but do not survive Benjamini--Hochberg correction at $q<0.05$ across
all 66 tests in the paper (also used in Table~\ref{tab:churn}).}
\label{tab:adv}
\setlength{\tabcolsep}{3.2pt}
\begin{tabular}{llcccc}
\toprule
Model & Quantizer & BF16 fooled \% & Quant. fooled \% & harm:help & $p$ \\
\midrule
Qwen3-0.6B & NF4 & 5.87 & 7.30 & 27:9 & \textbf{0.004} \\
 & INT4-RTN & 5.87 & 10.48 & 71:13 & \textbf{$<$0.001} \\
 & GPTQ & 5.87 & 6.98 & 24:10 & \textbf{0.024} \\
Llama-3.2-1B & NF4 & 6.83 & 7.30 & 20:14 & 0.39 \\
 & INT4-RTN & 6.83 & 7.54 & 26:17 & 0.22 \\
 & GPTQ & 6.83 & 7.62 & 30:20 & 0.20 \\
Qwen3-1.7B & NF4 & 5.16 & 6.03 & 14:3 & \textbf{0.013} \\
 & INT4-RTN & 5.16 & 6.98 & 35:12 & \textbf{0.001} \\
 & GPTQ & 5.16 & 5.40 & 10:7 & 0.63 \\
Llama-3.2-3B & NF4 & 6.03 & 6.27 & 12:9 & 0.66 \\
 & INT4-RTN & 6.03 & 6.19 & 14:12 & 0.85 \\
 & GPTQ & 6.03 & 6.35 & 10:6 & 0.45 \\
Qwen3-8B & NF4 & 4.37 & 4.44 & 6:5 & 1.00 \\
 & INT4-RTN & 4.37 & 4.60 & 10:7 & 0.63 \\
 & GPTQ & 4.37 & 4.21 & 5:7 & 0.77 \\
\textit{Pooled: Skywork-Reward-V2} & NF4 & -- & -- & 73:35 & \textbf{$<$0.001} \\
 & INT4-RTN & -- & -- & 146:54 & \textbf{$<$0.001} \\
 & GPTQ & -- & -- & 74:43 & \textbf{0.005} \\
\midrule
GRM-gemma2-2B & NF4 & 15.95 & 14.37 & 6:26 & \textbf{$<$0.001} \\
 & INT4-RTN & 15.95 & 16.51 & 28:21 & 0.39 \\
GRM-Llama3.2-3B & NF4 & 8.97 & 8.33 & 7:15 & 0.13 \\
 & INT4-RTN & 8.97 & 10.00 & 23:10 & \textbf{0.035}$^{\dagger}$ \\
\textit{Pooled: GRM} & NF4 & -- & -- & 13:41 & \textbf{$<$0.001} \\
 & INT4-RTN & -- & -- & 51:31 & \textbf{0.035}$^{\dagger}$ \\
\bottomrule
\end{tabular}

\end{table}

We report two negative results, both of which shaped the method above.

\paragraph{Learned scale tuning cannot beat \gpt{}.}
The model gives a crisp objective: minimize $\sig$, the decision-margin noise, rather than weight
reconstruction error. We implemented this directly. \emph{MarginTune} learns per-group INT4 scales
by straight-through estimation against a decision-aware objective, using the model's own BF16
margins as the teacher on unlabeled preference pairs (roughly $0.1\%$ of parameters, 300--800
steps).

Applied to \rtn{}, it works modestly: $\sig$ falls by $11$--$19\%$ and it recovers $25$--$35\%$ of
the quantization-induced adversarial fooling across four models. But applied to a \gpt{}
initialization it does nothing at all: $\sig$ changes by $-2\%$ to $+7\%$, which is noise, and
fooling rates are unchanged or slightly worse. We also could not cleanly separate the decision-aware
objective from a plain score-matching ablation, so we do not claim decision-awareness was the
active ingredient even in the \rtn{} case.

We read this as informative rather than merely disappointing. Within the space we searched,
straight-through tuning of per-group scales, \gpt{}'s residual $\sig$ could not be reduced,
which suggests it is not an artifact of scale calibration; we do not claim it is irreducible by
methods that optimize different quantities, such as learned clipping and equivalent
transformations \citep{shao2024omniquant} or rotations. If the noise cannot be tuned away, the remaining option is to
route around it, which is precisely what selective escalation does, and why it composes with
\gpt{} instead of competing with it.

\paragraph{Elite adversaries wash the effect out.}
We expected a stronger adversary to amplify the vulnerability, so we rebuilt the trap pools with a
32B adversary and selector-pressure filtering that kept only the top-3 most-attractive wrong
candidates. The effect got \emph{weaker}, not stronger: pooled over 1{,}540 problems, no model
reached significance ($p=0.71$, $0.21$, $0.22$ at 0.6B/1B/1.7B, and reversed at 3B).

The model explains this too. Traps only matter while they sit inside the vulnerability window. Elite
traps are so attractive that they beat the correct answer in BF16 as well: they leave the window
on the other side, and the BF16 model is fooled with or without quantization. Quantization harm is a
\emph{moderate-adversary} phenomenon, which is a meaningful restriction on the threat model and one
we would rather state than let a reviewer discover.

\end{document}